\documentclass[conference]{IEEEtran}
\IEEEoverridecommandlockouts
\usepackage{cite}
\usepackage{amsmath,amssymb,amsfonts}
\usepackage{algorithmic}
\usepackage{graphicx}
\usepackage{textcomp}
\usepackage{xcolor}
\usepackage{booktabs}
\usepackage{makecell}
\usepackage{lineno}

\makeatletter
\let\old@outputpage\@outputpage
\def\@outputpage{%
  \if@twocolumn
    \ifodd\count\z@ 
      \rightlinenumbers
    \else
      \leftlinenumbers
    \fi
  \fi
  \old@outputpage
}
\makeatother
\def\BibTeX{{\rm B\kern-.05em{\sc i\kern-.025em b}\kern-.08em
    T\kern-.1667em\lower.7ex\hbox{E}\kern-.125emX}}
\begin{document}

\title{GhostNetV3-Small: A Tailored Architecture and Comparative Study of Distillation Strategies for Tiny Images\\}

\author{\IEEEauthorblockN{Florian Zager}
\IEEEauthorblockA{\textit{ETIT-KIT} \\
Karlsruhe, Germany \\
}
\and
\IEEEauthorblockN{Hamza A. A. Gardi}
\IEEEauthorblockA{\textit{IIIT at ETIT- KIT} \\
Karlsruhe, Germany \\
}
}

\maketitle

\begin{abstract}
Deep neural networks have achieved remarkable success across a range of tasks, however their computational demands often make them unsuitable for deployment on resource-constrained edge devices. This paper explores strategies for compressing and adapting models to enable efficient inference in such environments. We focus on GhostNetV3, a state-of-the-art architecture for mobile applications, and propose GhostNetV3-Small, a modified variant designed to perform better on low-resolution inputs such as those in the CIFAR-10 dataset. In addition to architectural adaptation, we provide a comparative evaluation of knowledge distillation techniques, including traditional knowledge distillation, teacher assistants, and teacher ensembles. Experimental results show that GhostNetV3-Small significantly outperforms the original GhostNetV3 on CIFAR-10, achieving an accuracy of 93.94\%. Contrary to expectations, all examined distillation strategies led to reduced accuracy compared to baseline training. These findings indicate that architectural adaptation can be more impactful than distillation in small-scale image classification tasks, highlighting the need for further research on effective model design and advanced distillation techniques for low-resolution domains.
\footnote{Our code is available at https://github.com/FlorianREGAZ/PML}
\end{abstract}


\section{Introduction}
Deep neural networks \cite{hinton2006dl} have established themselves as the state-of-the-art across a wide range of domains, including computer vision \cite{huang2017densely}, speech recognition \cite{han2017capio}, and natural language processing \cite{devlin2019bert}. Their success is largely attributed to increasing model depth and parameter count, which typically correlates with improved accuracy. However, this growth in complexity comes at the cost of substantial computational and memory demands, which are often incompatible with the constraints of edge devices such as smartphones, embedded systems, and IoT hardware. As a result, there is a pressing need for techniques that allow smaller models to achieve performance levels comparable to their larger counterparts.

To address this challenge, several model compression methods have been proposed in recent years. Among them, knowledge distillation (KD) \cite{hinton2015distilling} has gained particular attention due to its conceptual simplicity and empirical effectiveness. In this framework, a large, pre-trained "teacher" network is used to guide the training of a smaller "student" model. Rather than learning solely from the hard labels of the training data, the student also learns from the soft predictions of the teacher, which contain additional information about class similarities. This is typically formalized through a loss function that combines standard cross-entropy with a Kullback–Leibler (KL) divergence term:

\begin{align}
\mathcal{L}_{\text{KD}} = \alpha \cdot \mathcal{L}_{\text{CE}}(y_{\text{true}}, y_{\text{S}}) + (1 - \alpha) \cdot T^2 \cdot \mathcal{L}_{\text{KL}}(y_{\text{T}}^T, y_{\text{S}}^T)
\end{align}

where $y_{\text{T}}^T$ and $y_{\text{S}}^T$ represent the soft logits of the teacher and student at temperature $T$, and $\alpha$ controls the trade-off between the hard and soft targets. Here, $y_{\text{true}}$ denotes the ground truth class labels, and $y_{\text{S}}$ refers to the student model’s predicted logits (or probabilities) used in the standard cross-entropy loss. The temperature $T$ is used to soften the output distributions by scaling the logits before applying the softmax function. Higher values of $T$ produce smoother probability distributions that reveal richer information about class similarities.

Despite its advantages, a known limitation of traditional KD is its reduced effectiveness when there is a large capacity gap between teacher and student networks. To mitigate this, the teacher assistant \cite{mirzadeh2020improved} strategy was introduced, where a sequence of intermediate-sized models is used to gradually transfer knowledge from the teacher to the student in a multi-step manner.

In parallel, ensemble teacher \cite{shen2019meal, fukuda2017efficient, you2017learning, liu2020adaptive} approaches have been proposed to enhance the richness of supervision. While the ensemble teacher method enhances supervision by introducing model diversity, it still falls short in addressing the capacity gap between large teachers and compact students. To effectively deploy knowledge distillation in real-world mobile scenarios, it is equally important to consider the architecture of the student model itself.

GhostNetV3 (GN-D) \cite{liu2024ghostnetv3} is one of the most efficient architectures currently available for mobile applications, striking a strong balance between accuracy and latency, that is also trained by leveraging KD. However, like many modern lightweight models, its architecture is optimized for high-resolution datasets such as ImageNet \cite{deng2009imagenet} (224×224 px images), which limits its effectiveness on smaller images like those in CIFAR-10 \cite{krizhevsky2009cifar10} (32×32 px images). This paper addresses this problem by proposing a modified architecture of GhostNetV3 tailored specifically for low-resolution inputs.

At the same time, we explore and compare different distillation strategies applied to both the original and the adapted GhostNetV3 architectures: standard training without distillation, original knowledge distillation, teacher assistant-based distillation, and ensemble teacher distillation.

In summary, our main contributions are the following:
\begin{itemize}
  \item We propose GhostNetV3-Small, with a new set of hyperparameters and architectural adjustments to the GhostNetV3 model, to reduce complexity and improve performance on smaller images.
  \item We provide a comprehensive overview of training techniques for GhostNetV3 and GhostNetV3-Small (GN-S) to achieve the best possible performance for edge devices.
\end{itemize}

\section{Related Work}

\subsection{Knowledge Distillation}
With the rapid growth of mobile computing, significant efforts have been dedicated to model compression techniques aimed at deploying efficient neural networks on resource-constrained devices. Model compression \cite{buciluǎ2006model, han2015deep, yu2018nisp} broadly encompasses methods such as pruning, quantization, and knowledge distillation. It aims to reduce model size and computational cost while preserving accuracy.

Knowledge Distillation \cite{hinton2015distilling}, a prominent subfield of model compression, enables a smaller student model to learn from the softened outputs of a larger teacher network, thereby transferring the teacher’s knowledge effectively.

After the introduction of this classical approach of mimicking teacher outputs, alternative Knowledge Distillation methods have been explored to improve knowledge transfer. For instance, FitNets \cite{romero2014fitnets} introduced the transfer of hidden activation outputs to guide the student’s intermediate representations. Attention transfer \cite{zagoruyko2016paying} methods convey spatial attention maps as knowledge, improving student network performance. Another approach views distilled knowledge as the flow of information between teacher network layers, which helps the student learn faster and generalize better, even across different tasks \cite{yim2017gift}.

More recent developments include self-distillation \cite{zhang2019your}, where a network improves itself by acting as its own teacher, and peer-based online distillation frameworks \cite{chen2020online}, where multiple student networks learn collaboratively with auxiliary peers and a group leader.

Knowledge distillation is highly relevant in modern compact model training. For example, GhostNetV3 \cite{liu2024ghostnetv3}, a state-of-the-art compact network optimized for mobile applications, achieves its strong performance largely by leveraging advanced knowledge distillation strategies during training.

\subsection{Teacher Assistant}

Teacher assistant \cite{mirzadeh2020improved} has been introduced to address the challenge of large capacity gaps between teacher and student models \cite{cho2019efficacy} in knowledge distillation. The paper proposed a multi-step distillation process, where an intermediate-sized teacher assistant network bridges the gap, enabling more effective knowledge transfer.

Subsequent studies have further enhanced this concept. For example, leveraging multiple teacher assistants, that are arranged in a decreasing size sequence to effectively bridge the capacity gap between teacher and student models. Each assistant guides both the next smaller assistant and the student, while stochastic teaching randomly drops some teachers during training to regularize learning. An example of this method would be Densely Guided Knowledge Distillation (DGKD) \cite{son2021densely}.

\subsection{Ensemble of Teachers}
Originally, knowledge distillation aimed to replace computationally expensive ensembles with a single efficient student model to reduce resource consumption during inference. However, recent advances leverage ensembles of teacher networks during training to transfer their combined knowledge into a compact student model, accepting the higher training cost for improved performance.

Ensembles provide richer and more diverse supervision, enabling the student network to learn knowledge more effectively. For instance, adversarial learning techniques have been applied to distill knowledge from multiple teacher models simultaneously \cite{shen2019meal}. For instance in acoustic modeling, ensembles of teachers have been successfully used to improve student performance \cite{fukuda2017efficient}.

Beyond output-level distillation, some approaches incorporate intermediate layer information from multiple teachers, imposing constraints to encourage diversity and better representation learning \cite{you2017learning}. Furthermore, adaptive frameworks like AMTML-KD \cite{liu2020adaptive} have been proposed that weigh the importance of each teacher differently depending on the input example, enabling more fine-grained knowledge transfer.

\subsection{Networks}
The development of lightweight CNNs for mobile applications has focused on progressively improving efficiency and accuracy. ShuffleNetV2 \cite{Ma_2018_ECCV_shufflenetv2} was introduced to bridge the gap between theoretical FLOPs reduction and actual speed on mobile devices by optimizing memory access and minimizing fragmentation. Building on this, MobileNetV3 \cite{koonce2021mobilenetv3} combined neural architecture search with manual design to incorporate efficient attention mechanisms and improved activation functions, further enhancing performance on mobile CPUs. GhostNetV1 \cite{han2020ghostnet} then addressed feature redundancy by generating additional “ghost” feature maps through inexpensive operations, significantly reducing computation while preserving accuracy. GhostNetV2 \cite{tang2022ghostnetv2} improved upon this by adding channel-wise attention and better feature fusion, boosting accuracy without increasing cost. The latest GhostNetV3 \cite{liu2024ghostnetv3} refined these concepts and introduced advanced training strategies, achieving superior accuracy-efficiency trade-offs optimized for modern mobile hardware. Each iteration has tackled specific limitations of prior models, steadily advancing compact CNN design for resource-constrained environments.

\section{Experimental Setup}
This section introduces the dataset employed for training and evaluation, followed by an overview of the neural network architectures utilized in our experiments. We then conclude with the implementation details.

\subsection{Dataset}
For all experiments, we used the CIFAR-10 \cite{krizhevsky2009cifar10} dataset, a widely adopted benchmark in image classification tasks. CIFAR-10 consists of 60,000 RGB images evenly distributed across 10 classes, with 50,000 images for training and 10,000 for testing. Each image has a resolution of 32×32 pixels.

\subsection{Networks}
For the student model, we use the default version of GhostNetV3, designed for high-resolution datasets like ImageNet. Given that CIFAR-10 images are significantly smaller, we also introduce a custom variant, GhostNetV3-Small, tailored to better accommodate low-resolution inputs. This smaller version maintains the core design principles of GhostNetV3 while reducing architectural complexity to avoid overfitting and unnecessary capacity. GhostNetV3-Small is being evaluated on different widths to find the optimal settings.

For the teacher models in our knowledge distillation experiments, we employ multiple well-established networks: ResNet-50, ResNet-18 \cite{he2016resnet}, VGG-13 \cite{simonyan2014vgg}, EfficientNetV2 \cite{tan2021efficientnetv2}, and the GhostNetV3-Small (2.8x). All of which have demonstrated strong performance on large-scale image classification tasks.

In the Teacher Assistant setup, we use progressive intermediate models to bridge the performance and capacity gap between the teacher and student networks. Specifically, we evaluate the sequence ResNet-50 $\rightarrow$ ResNet-34 $\rightarrow$ ResNet-18 $\rightarrow$ Student. With the students being GhostNetV3 and GhostNetV3-Small with different widths.

For the Ensemble Teacher setup, we construct an ensemble composed of four different architectures: DenseNet-161 \cite{huang2017densely}, VGG-13, ResNet-50, and InceptionV3 \cite{szegedy2016inception}.

\subsection{Implementation Details}
All experiments were implemented using the PyTorch deep learning framework, including preprocessing, model training, optimization, and evaluation routines.
For data preprocessing, we applied standard CIFAR-10 augmentation techniques: each input image was randomly cropped to 32×32 pixels with a padding of 4 pixels, followed by a random horizontal flip. The images were then normalized using the CIFAR-10 dataset-specific mean and standard deviation.
Model training was performed using stochastic gradient descent (SGD) with Nesterov momentum set to 0.9 and a weight decay of 0.01. The learning process spanned 200 epochs, beginning with an initial learning rate of 0.01. A linear warm-up was applied during the first 30\% of training, after which a cosine annealing learning rate schedule was used.
All models were trained with a batch size of 256, and all trainings and evaluations were conducted on a single NVIDIA RTX 3060 Ti GPU.

\subsection{Evaluation}
In all experiments, the model performance was evaluated using top-1 accuracy on the CIFAR-10 dataset, following its standard training/test split. During training, the model was evaluated on the test set after every epoch. The best top-1 accuracy achieved at any point during training was recorded as the model's final performance. All experiments were conducted using a single fixed random seed, and results were not averaged over multiple runs.

\section{Experimental Results}
\subsection{Model Baseline Performance}

The model baseline results in Table~\ref{tab:baseline} show that increasing the width of GhostNetV3-Small improves accuracy at the cost of model size. 

\begin{table}[ht]
\centering
\caption{Baseline performance of GhostNetV3 variants without distillation.}
\begin{tabular}{lccc}
\toprule
\textbf{Model} & \textbf{Width Multiplier} & \textbf{Params (M)} & \textbf{Top-1 Acc. (\%)} \\
\midrule
GN-S & 1.0x & 0.483 & 91.37 \\
GN-S & 1.3x & 0.712 & 92.35 \\
GN-S & 1.6x & 0.973 & 92.63 \\
GN-S & 1.9x & 1.277 & 93.23 \\
GN-S & 2.2x & 1.617 & 93.26 \\
GN-S & 2.8x & 2.426 & \textbf{93.94} \\
GN-S & 3.4x & 3.380 & 93.61 \\
GN-D & 1.0x & 6.860 & 91.23 \\
\midrule
ResNet18 & -- & 11.174 & 93.07 \\
ResNet34 & -- & 21.282 & 93.34 \\
ResNet50 & -- & 23.521 & 93.65 \\
VGG13 & -- & 28.334 & 94.22 \\
InceptionV3 & -- & 21.640 & 93.74 \\
DensNet161 & -- & 26.483 & 94.07 \\
EfficientNetV2 & -- & 20.190 & \textbf{97.90} \\
\bottomrule
\end{tabular}
\label{tab:baseline}
\end{table}

\begin{figure}[htbp]
  \centering
  \includegraphics[width=0.4\textwidth]{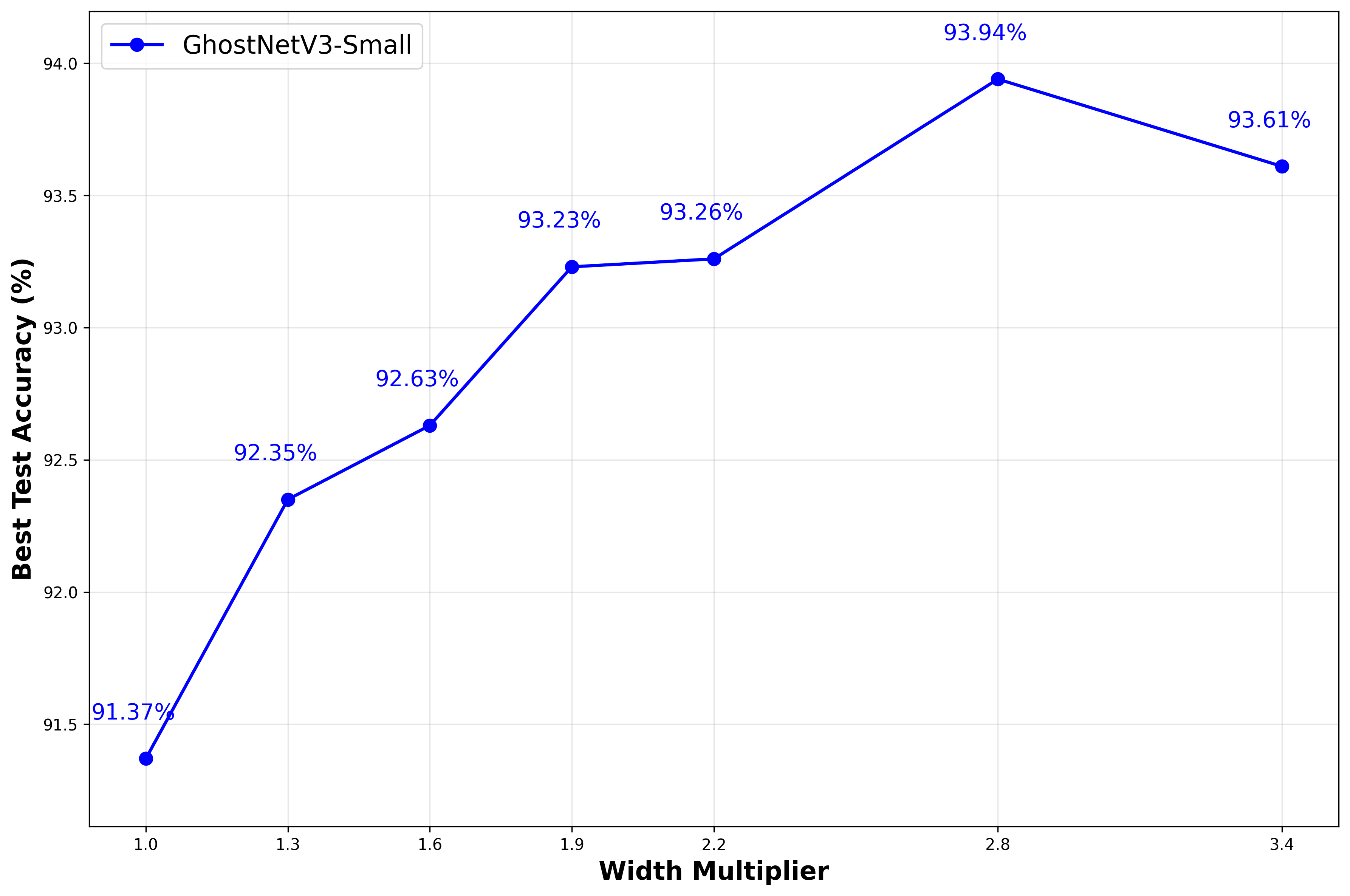}
  \caption{The top-1 accuracy achieved with various width multipliers.}
  \label{fig:gn3s_width_acc}
\end{figure}

These baselines serve as the reference point for evaluating the benefits of various knowledge distillation strategies. As shown in Fig. \ref{fig:gn3s_width_acc}, the top-1 accuracy of GhostNetV3-Small peaks at 93.94\% with the 2.8x configuration. Beyond this point, accuracy slightly declines, as observed at the 3.4x setting. Importantly, all GhostNetV3-Small variants consistently outperform the default GhostNetV3 model, despite having up to ten times fewer parameters.

\subsection{Effect of Different Teachers on Distillation}
As shown in Table~\ref{tab:kd_teachers}, all distillation results yield negative deltas compared to their respective baselines, indicating that knowledge distillation in these cases degraded the student model's performance. 

\begin{table}[ht]
\centering
\caption{Knowledge Distillation performance.}
\begin{tabular}{lccc}
\toprule
\textbf{Student Model} & \textbf{Teacher Model} & \textbf{Top-1 Acc. (\%)} & \textbf{$\Delta$ vs. Baseline} \\
\midrule
GN-S (1.0x) & ResNet18 & 91.10 & -0.27 \\
GN-S (1.0x) & ResNet50 & 91.21 & -0.16 \\
GN-S (1.0x) & VGG13 & 91.19 & -0.18 \\
GN-S (1.0x) & GN-S (2.8x) & 90.90 & -0.47 \\
GN-S (2.8x) & VGG13 & 93.58 & -0.36 \\
GN-S (2.8x) & EfficientNetV2 & 92.16 & -1.78 \\
GN-D (1.0x) & ResNet50 & 91.10 & -0.13 \\
GN-D (1.0x) & VGG13 & 90.93 & -0.3 \\
\bottomrule
\end{tabular}
\label{tab:kd_teachers}
\end{table}

A possible explanation for this outcome is that the teacher models may have been too large or complex for the student model to effectively learn from. However, this hypothesis is challenged by the fact that one of the worst-performing pairs is GhostNetV3-Small (1.0x) as the student and GhostNetV3-Small (2.8x) as the teacher. Despite having the smallest gap in model size and architecture complexity, this configuration leads to the most significant drop in accuracy. This suggests that model compatibility and teacher quality are not solely determined by size, and more nuanced factors may be at play.

The largest observed decrease in performance occurred when using EfficientNetV2 as the teacher and GhostNetV3-Small (2.8x) as the student. Although EfficientNetV2 was pretrained on ImageNet and subsequently fine-tuned on CIFAR-10, its architecture and training rely on high-resolution inputs (224×224 px). This mismatch with the student’s lower-resolution input space (32×32 px) may have hindered effective knowledge transfer. Despite the teacher model achieving strong standalone performance, it failed to convey that knowledge in a beneficial way to the student. This indicates a required compatibility not just in model size, but also in input resolution and representational alignment.

\subsection{Teacher Assistant Distillation}

The teacher assistant strategy was applied using a sequential distillation path: ResNet50 $\rightarrow$ ResNet34 $\rightarrow$ ResNet18. 

Despite the intention of easing the learning process by gradually reducing teacher complexity, the results in Table~\ref{tab:ta_table} show a negative impact on performance compared to the baseline. This mirrors the behavior observed in direct knowledge distillation, where all delta values were also negative.

\begin{table}[ht]
\centering
\caption{
Teacher Assistant performance.}
\begin{tabular}{lccc}
\toprule
\textbf{Student Model} & \textbf{Top-1 Acc. (\%)} & \textbf{$\Delta$ vs. Baseline} \\
\midrule
GN-S (1.0x) & 90.60 & -0.77 \\
GN-S (1.3x) & 91.85 & -0.50 \\
GN-S (1.6x) & 92.50 & -0.13 \\
GN-D (1.0x) & 90.66 & -0.57 \\
\bottomrule
\end{tabular}
\label{tab:ta_table}
\end{table}

This outcome aligns with expectations, considering that ResNet18 previously failed to improve the student model when used as a standalone teacher. In the teacher assistant setup, the final distillation stage still relies on ResNet18, meaning the effective complexity of the final teacher remains unchanged. Therefore, it is unsurprising that performance did not improve, underlining the importance of selecting effective teacher models for successful distillation.

\subsection{Teacher Ensemble Distillation}
To further explore the impact of different knowledge distillation strategies, a teacher ensemble was constructed using four diverse architectures: ResNet50, DenseNet161, VGG13, and InceptionV3. The outputs of these models were averaged to generate the soft targets for the student model, following the standard ensemble distillation approach.

As shown in Table~\ref{tab:ensemble_table}, all ensemble-based distillation experiments resulted in negative accuracy deltas compared to their respective baselines. 

\begin{table}[ht]
\centering
\caption{
Teacher Ensemble performance.}
\begin{tabular}{lccc}
\toprule
\textbf{Student Model} & \textbf{Top-1 Acc. (\%)} & \textbf{$\Delta$ vs. Baseline} \\
\midrule
GN-S (1.0x) & 90.51 & -0.86 \\
GN-S (2.8x) & 93.28 & -0.66 \\
GN-D (1.0x) & 90.83 & -0.40 \\
\bottomrule
\end{tabular}
\label{tab:ensemble_table}
\end{table}

Notably, the degradation in accuracy was in most cases even more pronounced than with direct distillation or teacher assistant setups. This suggests that increasing teacher diversity alone is not sufficient to yield better student performance, particularly in the context of compact models trained on low-resolution datasets.

\section{Conclusion and Outlook}

In this paper, we introduced GhostNetV3-Small, a modified version of the GhostNetV3 architecture designed specifically for low-resolution image inputs. On the CIFAR-10 dataset, our adapted model significantly outperforms the original GhostNetV3.

Additionally, we evaluated several knowledge distillations like traditional knowledge distillation, teacher assistant, and teacher ensemble in combination with GhostNetV3-Small and the default GhostNetV3. Interestingly, none of these approaches surpassed the performance of standard training without distillation in our experimental setup.

Looking ahead, more advanced knowledge distillation techniques, such as AMTML-KD or DGKD, may offer improved performance in this context and warrant further investigation. Future work could also explore a wider variety of teacher models, including those based on transformer architectures, alternative assistant sequences, and more diverse teacher ensembles. Finally, while this study focused exclusively on CIFAR-10, evaluating the proposed methods on other datasets will be crucial for assessing generalizability and practical applicability.


\begin{thebibliography}{10}

\bibitem{buciluǎ2006model}
Cristian Buciluǎ, Rich Caruana, and Alexandru Niculescu-Mizil.
\newblock Model compression.
\newblock In {\em Proceedings of the 12th ACM SIGKDD international conference on Knowledge discovery and data mining}, pages 535--541, 2006.

\bibitem{chen2020online}
Defang Chen, Jian-Ping Mei, Can Wang, Yan Feng, and Chun Chen.
\newblock Online knowledge distillation with diverse peers.
\newblock In {\em Proceedings of the AAAI conference on artificial intelligence}, volume~34, pages 3430--3437, 2020.

\bibitem{cho2019efficacy}
Jang~Hyun Cho and Bharath Hariharan.
\newblock On the efficacy of knowledge distillation.
\newblock In {\em Proceedings of the IEEE/CVF international conference on computer vision}, pages 4794--4802, 2019.

\bibitem{deng2009imagenet}
Jia Deng, Wei Dong, Richard Socher, Li-Jia Li, Kai Li, and Li~Fei-Fei.
\newblock Imagenet: A large-scale hierarchical image database.
\newblock In {\em 2009 IEEE conference on computer vision and pattern recognition}, pages 248--255. Ieee, 2009.

\bibitem{devlin2019bert}
Jacob Devlin, Ming-Wei Chang, Kenton Lee, and Kristina Toutanova.
\newblock Bert: Pre-training of deep bidirectional transformers for language understanding.
\newblock In {\em Proceedings of the 2019 conference of the North American chapter of the association for computational linguistics: human language technologies, volume 1 (long and short papers)}, pages 4171--4186, 2019.

\bibitem{fukuda2017efficient}
Takashi Fukuda, Masayuki Suzuki, Gakuto Kurata, Samuel Thomas, Jia Cui, and Bhuvana Ramabhadran.
\newblock Efficient knowledge distillation from an ensemble of teachers.
\newblock In {\em Interspeech}, pages 3697--3701, 2017.

\bibitem{han2020ghostnet}
Kai Han, Yunhe Wang, Qi~Tian, Jianyuan Guo, Chunjing Xu, and Chang Xu.
\newblock Ghostnet: More features from cheap operations.
\newblock In {\em Proceedings of the IEEE/CVF conference on computer vision and pattern recognition}, pages 1580--1589, 2020.

\bibitem{han2017capio}
Kyu~J Han, Akshay Chandrashekaran, Jungsuk Kim, and Ian Lane.
\newblock The capio 2017 conversational speech recognition system.
\newblock {\em arXiv preprint arXiv:1801.00059}, 2017.

\bibitem{han2015deep}
Song Han, Huizi Mao, and William~J Dally.
\newblock Deep compression: Compressing deep neural networks with pruning, trained quantization and huffman coding.
\newblock {\em arXiv preprint arXiv:1510.00149}, 2015.

\bibitem{he2016resnet}
Kaiming He, Xiangyu Zhang, Shaoqing Ren, and Jian Sun.
\newblock Deep residual learning for image recognition.
\newblock In {\em Proceedings of the IEEE conference on computer vision and pattern recognition}, pages 770--778, 2016.

\bibitem{hinton2015distilling}
Geoffrey Hinton, Oriol Vinyals, and Jeff Dean.
\newblock Distilling the knowledge in a neural network.
\newblock {\em arXiv preprint arXiv:1503.02531}, 2015.

\bibitem{hinton2006dl}
Geoffrey~E Hinton, Simon Osindero, and Yee-Whye Teh.
\newblock A fast learning algorithm for deep belief nets.
\newblock {\em Neural computation}, 18(7):1527--1554, 2006.

\bibitem{huang2017densely}
Gao Huang, Zhuang Liu, Laurens Van Der~Maaten, and Kilian~Q Weinberger.
\newblock Densely connected convolutional networks.
\newblock In {\em Proceedings of the IEEE conference on computer vision and pattern recognition}, pages 4700--4708, 2017.

\bibitem{koonce2021mobilenetv3}
Brett Koonce.
\newblock Mobilenetv3.
\newblock In {\em Convolutional neural networks with swift for tensorflow: image recognition and dataset categorization}, pages 125--144. Springer, 2021.

\bibitem{krizhevsky2009cifar10}
Alex Krizhevsky, Geoffrey Hinton, et~al.
\newblock Learning multiple layers of features from tiny images.
\newblock 2009.

\bibitem{liu2020adaptive}
Yuang Liu, Wei Zhang, and Jun Wang.
\newblock Adaptive multi-teacher multi-level knowledge distillation.
\newblock {\em Neurocomputing}, 415:106--113, 2020.

\bibitem{liu2024ghostnetv3}
Zhenhua Liu, Zhiwei Hao, Kai Han, Yehui Tang, and Yunhe Wang.
\newblock Ghostnetv3: Exploring the training strategies for compact models.
\newblock {\em arXiv preprint arXiv:2404.11202}, 2024.

\bibitem{Ma_2018_ECCV_shufflenetv2}
Ningning Ma, Xiangyu Zhang, Hai-Tao Zheng, and Jian Sun.
\newblock Shufflenet v2: Practical guidelines for efficient cnn architecture design.
\newblock In {\em Proceedings of the European Conference on Computer Vision (ECCV)}, September 2018.

\bibitem{mirzadeh2020improved}
Seyed~Iman Mirzadeh, Mehrdad Farajtabar, Ang Li, Nir Levine, Akihiro Matsukawa, and Hassan Ghasemzadeh.
\newblock Improved knowledge distillation via teacher assistant.
\newblock In {\em Proceedings of the AAAI conference on artificial intelligence}, volume~34, pages 5191--5198, 2020.

\bibitem{romero2014fitnets}
Adriana Romero, Nicolas Ballas, Samira~Ebrahimi Kahou, Antoine Chassang, Carlo Gatta, and Yoshua Bengio.
\newblock Fitnets: Hints for thin deep nets.
\newblock {\em arXiv preprint arXiv:1412.6550}, 2014.

\bibitem{shen2019meal}
Zhiqiang Shen, Zhankui He, and Xiangyang Xue.
\newblock Meal: Multi-model ensemble via adversarial learning.
\newblock In {\em Proceedings of the AAAI conference on artificial intelligence}, volume~33, pages 4886--4893, 2019.

\bibitem{simonyan2014vgg}
Karen Simonyan and Andrew Zisserman.
\newblock Very deep convolutional networks for large-scale image recognition.
\newblock {\em arXiv preprint arXiv:1409.1556}, 2014.

\bibitem{son2021densely}
Wonchul Son, Jaemin Na, Junyong Choi, and Wonjun Hwang.
\newblock Densely guided knowledge distillation using multiple teacher assistants.
\newblock In {\em Proceedings of the IEEE/CVF International Conference on Computer Vision}, pages 9395--9404, 2021.

\bibitem{szegedy2016inception}
Christian Szegedy, Vincent Vanhoucke, Sergey Ioffe, Jon Shlens, and Zbigniew Wojna.
\newblock Rethinking the inception architecture for computer vision.
\newblock In {\em Proceedings of the IEEE conference on computer vision and pattern recognition}, pages 2818--2826, 2016.

\bibitem{tan2021efficientnetv2}
Mingxing Tan and Quoc Le.
\newblock Efficientnetv2: Smaller models and faster training.
\newblock In {\em International conference on machine learning}, pages 10096--10106. PMLR, 2021.

\bibitem{tang2022ghostnetv2}
Yehui Tang, Kai Han, Jianyuan Guo, Chang Xu, Chao Xu, and Yunhe Wang.
\newblock Ghostnetv2: Enhance cheap operation with long-range attention.
\newblock {\em Advances in Neural Information Processing Systems}, 35:9969--9982, 2022.

\bibitem{yim2017gift}
Junho Yim, Donggyu Joo, Jihoon Bae, and Junmo Kim.
\newblock A gift from knowledge distillation: Fast optimization, network minimization and transfer learning.
\newblock In {\em Proceedings of the IEEE conference on computer vision and pattern recognition}, pages 4133--4141, 2017.

\bibitem{you2017learning}
Shan You, Chang Xu, Chao Xu, and Dacheng Tao.
\newblock Learning from multiple teacher networks.
\newblock In {\em Proceedings of the 23rd ACM SIGKDD international conference on knowledge discovery and data mining}, pages 1285--1294, 2017.

\bibitem{yu2018nisp}
Ruichi Yu, Ang Li, Chun-Fu Chen, Jui-Hsin Lai, Vlad~I Morariu, Xintong Han, Mingfei Gao, Ching-Yung Lin, and Larry~S Davis.
\newblock Nisp: Pruning networks using neuron importance score propagation.
\newblock In {\em Proceedings of the IEEE conference on computer vision and pattern recognition}, pages 9194--9203, 2018.

\bibitem{zagoruyko2016paying}
Sergey Zagoruyko and Nikos Komodakis.
\newblock Paying more attention to attention: Improving the performance of convolutional neural networks via attention transfer.
\newblock {\em arXiv preprint arXiv:1612.03928}, 2016.

\bibitem{zhang2019your}
Linfeng Zhang, Jiebo Song, Anni Gao, Jingwei Chen, Chenglong Bao, and Kaisheng Ma.
\newblock Be your own teacher: Improve the performance of convolutional neural networks via self distillation.
\newblock In {\em Proceedings of the IEEE/CVF international conference on computer vision}, pages 3713--3722, 2019.

\end{thebibliography}

\end{document}